\documentclass[runningheads]{llncs}
\usepackage{url}
\usepackage{bm}
\usepackage{gensymb}
\usepackage{makecell}
\usepackage{cite}
\usepackage{amsmath,amssymb,amsfonts,enumitem}
\usepackage[noend]{algorithmic}
\usepackage{algorithm}
\usepackage{etoolbox}
\usepackage{textcomp}
\usepackage{xcolor}
\usepackage{color, colortbl}
\usepackage{booktabs}
\usepackage{tabularx}
\usepackage{tabulary}
\usepackage{adjustbox}
\usepackage{xspace}
\usepackage{float}
\usepackage{graphicx}
\usepackage{tabularx}
\usepackage{wrapfig}
\usepackage{footnote}
\usepackage{comment}
\usepackage{multirow}
\usepackage{array}
\usepackage{capt-of}
\usepackage{varwidth}
\usepackage[utf8]{inputenc}
\usepackage{nccmath}
\usepackage[caption=false]{subfig}
\usepackage[]{hyperref}
\usepackage{bbding}

\definecolor{Gray}{gray}{0.9}
\usepackage[T1]{fontenc}
\usepackage{graphicx}
\makeatletter
\def\thanks#1{\protected@xdef\@thanks{\@thanks
        \protect\footnotetext{#1}}}
\makeatother

\begin{document}
\title{EdgeMA: Model Adaptation System for Real-Time Video Analytics on Edge Devices}
%
%
\author{Liang Wang$^{1,*}$, Nan Zhang$^{2,*}$\thanks{\hspace{-1em} * These two authors have contributed to this work equally.}, Xiaoyang Qu\inst{2,}\small{\Envelope}\thanks{\hspace{-1em} \small{\Envelope} Corresponding author: Xiaoyang Qu (e-mail: quxiaoy@gmail.com).}, Jianzong Wang$^2$, Jiguang Wan$^1$, Guokuan Li$^1$, Kaiyu Hu$^3$, Guilin Jiang$^4$, Jing Xiao$^2$}
\authorrunning{L. Wang et al.}
\titlerunning{EdgeMA: Real-Time Video Analytics on Edge Devices}
%
\institute{
Huazhong University of Science and Technology, Wuhan, China \and
Ping An Technology (Shenzhen) Co., Ltd., Shenzhen, China
\and
Stony Brook University, Stony Brook, NY \and
Hunan Chasing Financial Holdings Co., Ltd., Changsha, China
}
\maketitle              
\begin{abstract}
Real-time video analytics on edge devices for changing scenes remains a difficult task. As edge devices are usually resource-constrained, edge deep neural networks (DNNs) have fewer weights and shallower architectures than general DNNs. As a result, they only perform well in limited scenarios and are sensitive to data drift. In this paper, we introduce EdgeMA, a practical and efficient video analytics system designed to adapt models to shifts in real-world video streams over time, addressing the data drift problem. EdgeMA extracts the gray level co-occurrence matrix based statistical texture feature and uses the Random Forest classifier to detect the domain shift. Moreover, we have incorporated a method of model adaptation based on importance weighting, specifically designed to update models to cope with the label distribution shift. Through rigorous evaluation of EdgeMA on a real-world dataset, our results illustrate that EdgeMA significantly improves inference accuracy.

\end{abstract}
\section{Introduction}
Real-time video analytics has become a promising application in the field of computer vision, which is powered by deep neural network (DNN) models, e.g., ResNet\cite{he2016deep} and EfficientNet\cite{tan2019efficientnet}. Video analytics applications such as traffic monitoring\cite{liao2023combining} use local cameras that continuously generate high-quality video streams to understand the environment. Most of these applications have to be carried out with real-time feedback. Therefore, edge computing is favored in video analytics because it eliminates the need for costly network overhead associated with uploading videos to the cloud and also reduces latency.

\begin{figure}[htbp]
    \centering
    \subfloat[Video snapshot (day)\label{fig1-1a}]{%
        \includegraphics[width=0.45\linewidth]{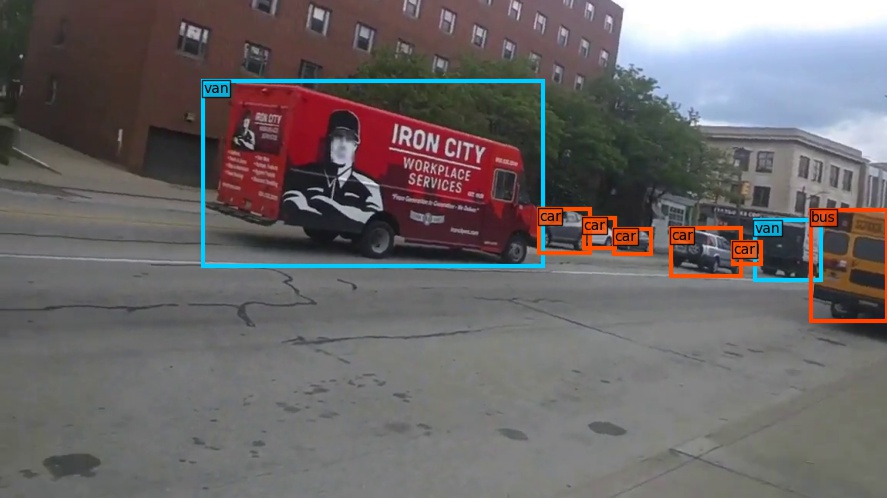}}
    \hfill
    \subfloat[Video snapshot (night)\label{fig1-1b}]{%
        \includegraphics[width=0.45\linewidth]{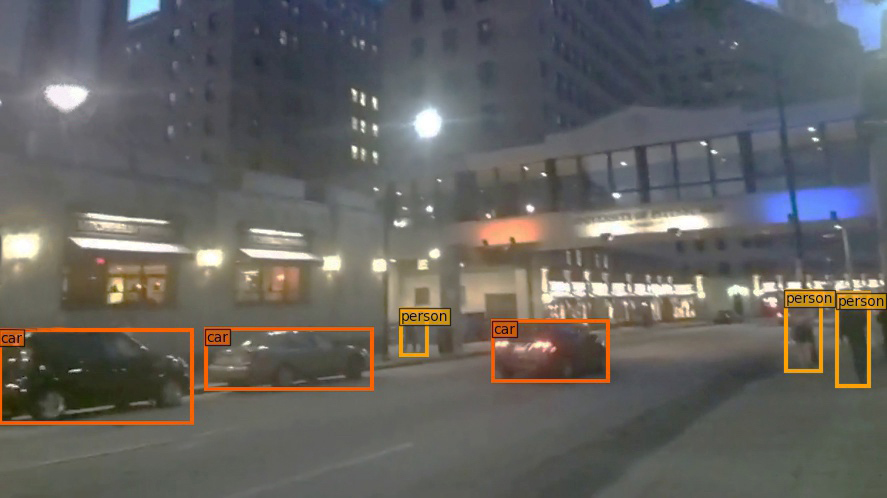}}
    \\
    \subfloat[Domain shift\label{fig1-1c}]{%
        \includegraphics[width=0.45\linewidth]{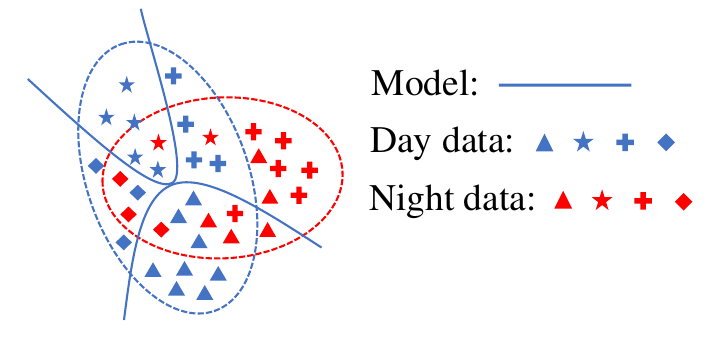}}
    \hfill
    \subfloat[Label distribution shift\label{fig1-1d}]{%
        \includegraphics[width=0.45\linewidth]{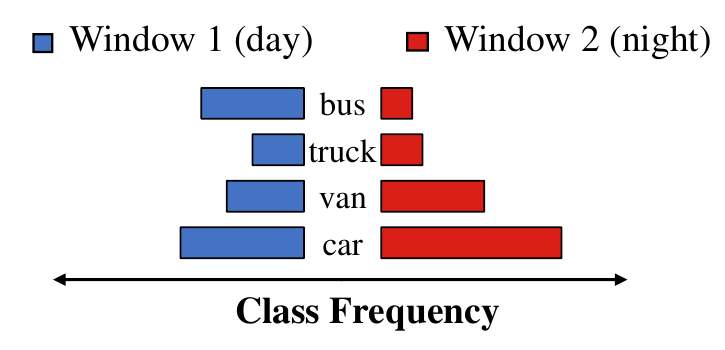}}
    \caption{(a) and (b) show the crossroad traffic in the video during the day and night, respectively.
    (c) shows the domains from day to night and the results of inference labels for data.
    We count the frequency of objects within each time window, and (d) shows the statistics results of class label distribution.}
    \label{fig1}
\end{figure}

Due to limited computing resources, edge devices typically use lightweight DNN models \cite{zhang2018shufflenet, howard2017mobilenets} for video analytics. Lightweight models have fewer weights and shallower architectures, often making them unsuitable for delivering high accuracy. In addition, when employed in the actual field, they are vulnerable to the data drift \cite{moreno2012unifying} problem: a mismatch between initial training data and live video data. Data drift results in substantial declines in accuracy as it violates the underlying assumption of DNN models that the training data from the past should resemble the test data in the future. This further aggravates the challenge of real-time video analytics.

In video analytics serving, domain and label distribution shifts are typical data drift phenomena \cite{jia2023counterfactual, khani2021real, bhardwaj2022ekya, liu2023fedet, khani2023recl}. As shown in Figure \ref{fig1}, street cameras encounter varying scenes over time. The label distribution of video varies over time, reducing the edge model’s accuracy. In addition, the domain goes from day to night. While the model trained on the data collected based on daytime conditions fails to work well when deployed in the dark, degrading the accuracy.

Having a static model throughout the entire life-long inference process often negatively impacts performance, particularly with edge models. Due to their constrained capacity to learn object variations, edge DNNs require regular updates to accommodate changing object distributions and ensure optimal accuracy. We must consider the migration from domain to domain (e.g., weather, light). 

In this work, we propose EdgeMA (\textbf{Edge} \textbf{M}odel \textbf{A}dapter), a video analytics system that resolves data drift in real-time video analytics on edge devices. EdgeMA copes with changes in the image domain and the label distribution over time, providing the model adaptation for edge computing. In summary, the main contributions of our work are the following:

\begin{itemize}
    \item We propose a novel system EdgeMA for real-time video analytics on edge devices to cope with both domain shift and label distribution shift.
    \item We design the lightweight domain detector to effectively identify and address domain shift.
    \item We detect label distribution shift and allow for real-time model retraining while ensuring the overall system remains resource-efficient.
    \item In our experiment on a real-world dataset, EdgeMA substantially outperforms scenarios without automatic model adaptation.
\end{itemize}

\section{Related Work}

\textbf{Video analytics systems.} Existing work has contributed to the creation of efficient video analytics systems. VideoStorm\cite{zhang2017live} investigates quality-lag requirements in video queries. Focus\cite{hsieh2018focus} uses low-cost models to index videos. Chameleon\cite{jiang2018chameleon} exploits correlations in camera content to amortize profiling costs.
NoScope\cite{kang2017noscope} enhances inference speed through cascading models and filtering mechanisms. 
Clownfish\cite{nigade2020clownfish} amalgamates inference outcomes from optimized models on edge devices with delayed results from comprehensive models in the cloud. 
However, these systems predominantly employ offline-trained models, overlooking the potential impact of model adaptation and data drift on accuracy.

\noindent \textbf{Edge computing.} The proliferation of the Internet of Things (IoT) has driven the production and usage of diverse hardware devices/sensors across the globe. These devices are able to collect data and then send it to the server for storage or processing, allowing end-users to access and extract the information as per their requirements \cite{shi2016promise}. Nevertheless, cloud computing has begun to reveal some issues. The centralized nature of cloud computing, handling data generated by global end devices, leads to numerous challenges, such as reduced throughput, increased latency, bandwidth constraints, data privacy concerns, and augmented costs. These challenges become particularly pressing in IoT applications, which demand rapid and low-latency data processing, analytics, and result delivery \cite{qin2018power, qu2020quantization}. To combat the aforementioned challenges associated with cloud computing, a novel computing paradigm known as edge computing, has attracted widespread attention \cite{hua2023edge}. In essence, edge computing offloads data processing, storage, and computing tasks traditionally assigned to the cloud to the network's edge, in close proximity to the terminal devices. This transition helps minimize data transmission and device response times, alleviate network bandwidth pressure, decrease data transmission costs, and facilitate a decentralized system \cite{ghosh2020edge}.

\noindent \textbf{Data drift.} Edge DNNs can only memorize a limited number of object appearances and scenarios. Hence, they are particularly vulnerable to data drift\cite{moreno2012unifying, jia2023counterfactual}, which arises when real-time video data significantly diverges across domains.
Variations of scene density (for instance, during rush hour) and lighting conditions (such as daytime versus nighttime) over time pose challenges for surveillance cameras attempting precise object detection. 
Additionally, the distribution of object classes changes over time, leading to a decrease in the precision of the edge model\cite{wang2023shoggoth}. Because of their restricted ability to memorize changes, edge models necessitate continuous retraining with the most recent data and shifting object distributions to maintain high accuracy.
Video-analytics systems are beginning to adopt continuous learning to adapt to changing video scenes and improve inference accuracy\cite{khani2021real, bhardwaj2022ekya, khani2023recl}. Ekya\cite{bhardwaj2022ekya} provides sophisticated resource-sharing mechanisms for efficient model retraining. AMS \cite{khani2021real} dynamically adjusts the frame sampling rate on edge devices depending on scene changes, mitigating the need for frequent retraining. RECL \cite{khani2023recl} combines Ekya and AMS, and offers more rapid responses by choosing an appropriate model from a repository of historical models during analysis. EdgeMA further considers lightweight retraining processing at the edge.

\section{System Design}

\begin{figure*}[htbp]
 \centering
 \includegraphics[width=1\textwidth]{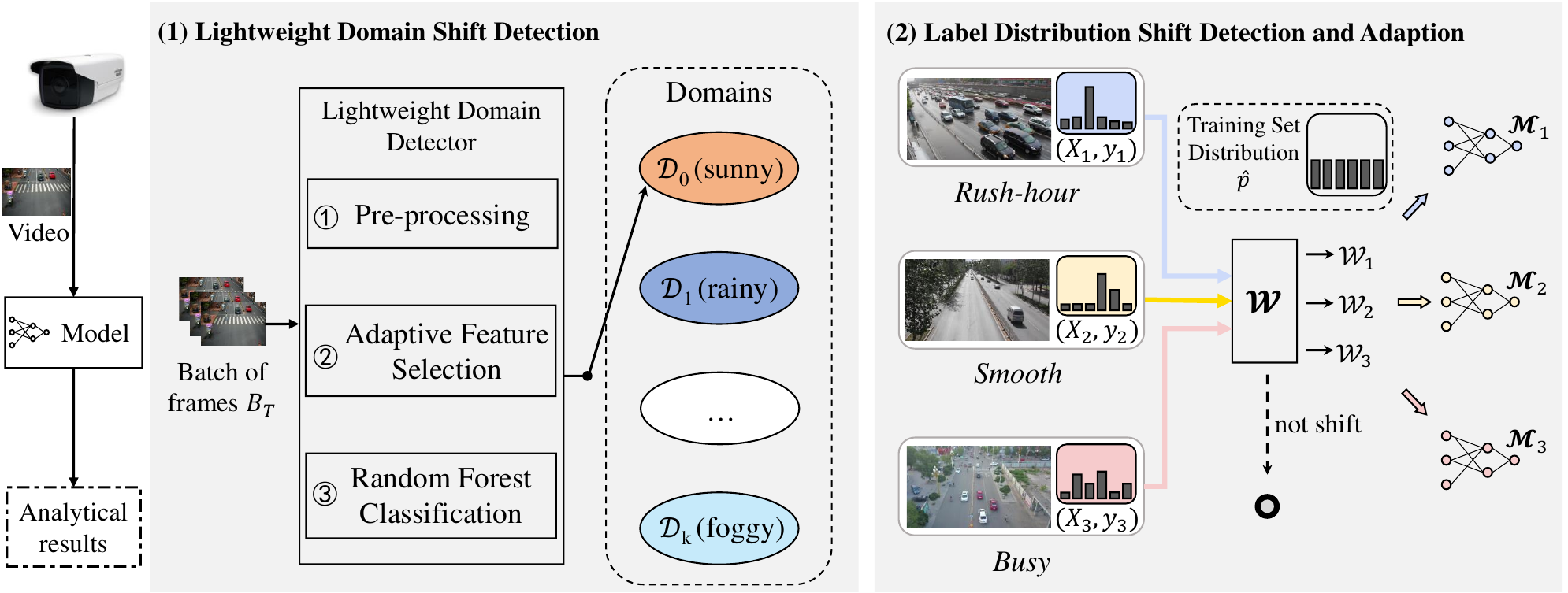}
 \caption{EdgeMA system overview.}
 \label{fig2}
\end{figure*}

Figure \ref{fig2} presents an overview of EdgeMA. Each edge device continuously performs inference for video analytics and buffers sampled video frames over time. The EdgeMA at the edge device runs iteratively on each new buffered batch of frames $B_T$, deciding whether to adapt the lightweight specialized model. The fundamental information upon which our system depends, when analyzing video feeds from static cameras or any other videos where the camera remains stationary, is that their domains and label distributions exhibit spatio-temporal locality. The model adaptation algorithm consists of two phases: (1) lightweight domain shift detection and (2) label distribution shift detection and adaptation.

In the first phase, EdgeMA detects in which domain the current batch lies. In our system, each domain corresponds to a training set containing a few data samples (e.g., 2000 images) from the domain scenario and an existing model pre-trained on a general dataset. After detecting, if the domain has changed, the model is fine-tuned using the corresponding training set to adapt to the specialized domain.

In the second phase, the edge device detects if the label distribution shifts. If the label distribution changes, the model also requires retraining with the dataset of the current domain. We assume that it makes sense to give more importance to the classes of training data whose frequency of appearance is nearest to the features of the live data to improve accuracy for different label distributions.

When our training set is small, retraining can perform well with low computing costs\cite{gu2020efficient}. And for concreteness, we describe our design for object detection, but the system is general and can be adapted to other tasks. Next, we illustrate the two phases of model adaptation in detail.

\subsection{Lightweight Domain Shift Detection}

The variations of illumination, weather, or style in different domains can seriously affect the model performance. Cross-domain detection for video is the challenge that model adaptation must overcome. Previous works\cite{xie2019multi, ibrahim2019weathernet} can use heavily computation-intensive networks for domain shift detection. However, edge devices must allocate computing resources as much as possible to the video analytics model. It is not desirable to run the video analytics model with the complex domain detection network simultaneously on a resource-constrained edge device. Therefore, we design a lightweight domain detector for EdgeMA. We detect the video's environment condition by a random forest classifier\cite{cutler2012random} (RFC) and Gray Level Co-occurrence Matrix (GLCM)-based texture features, to detect the video's environmental conditions. This approach not only ensures high accuracy but also minimizes computational expenses and conserves energy.

\textbf{Pre-processing Operation}. EdgeMA requires conversion of the frame to grayscale, wherein each pixel value represents spectral intensity ranging between 0 and 255. The grayscale image, represented by $g$, is derived by computing the luminance of the color image using equation (\ref{e1}):

\begin{equation}
g=0.299 \times R+0.587 \times G+0.114 \times B   \label{e1}
\end{equation}

\textbf{Adaptive Feature Selection}. 
For image classification, texture features, dependent on the repetitive patterns across an image region, are predominantly utilized. The GLCM of an image $I$ of dimensions $n*m$ characterizes the texture by quantifying occurrences of pixels with specific absolute values based on a spatial offset, is defined as:

\begin{equation}
C(g_i,\!g_j)\!=\!\sum_{p=1}^n\!\sum_{q=1}^m\!\begin{cases}1, \text{if } I(p,\!q)\!=\!g_i \text{ and } I(p\!+\!\Delta_x,\!q\!+\!\Delta_y)\!=\!g_j \\ 0, \text{otherwise}\end{cases}  \label{e2}
\end{equation}

\noindent where $g_i$ and $g_j$ represent the gray level values of an image. For spatial positions $p$ and $q$ within the image $I$, the offset $(\Delta_x, \Delta_y)$ is determined by both the direction and distance for which the matrix is computed.
GLCM features are computed with directions $\theta \in$ \{0\degree, 45\degree, 90\degree, 135\degree\}, distance $d \in$ [1, 30], and the 6 prevailing texture properties\cite{maji2023rvfl}, including contrast, correlation, homogeneity, angular second moment, dissimilarity, and energy. In total, 4*30*6=720 features are derived from GLCM.

Moreover, we employ the AdaBoost method\cite{rojas2009adaboost} for feature selection to ascertain the relative importance of features. AdaBoost not only selects the most significant features but also assigns weights to weak classifiers to enhance classification performance. The algorithm based on AdaBoost starts by choosing a feature as an initial weak learner. During each iteration $t$, the primary steps of the selection process are shown as follows. (1) Normalize the weights of the images as $w^{[t]}=\left\{w_I^{[t]}\right\}_I$, where $I$ is from the training image set.
(2) Choose the feature $F_j$ that yields the least classification error summed over all images, weighted by $w^{[t]}$.
(3) Increase the value $S_j$, denoting the importance score of the feature $F_j$.
(4) Adjust the weights for each image sample $I (w^{[t]})$. The weights should correlate with the error rate \(E\) for that image.
Through this method, images correctly classified by the chosen feature have diminished influence, whereas the weights of misclassified images increase. The final output of the algorithm is the vector $S$, signifying the relative importance of the original features.

\textbf{Random Forest Classification}. 
RFC\cite{cutler2012random} is a prominent method in machine learning, particularly for high-dimensional classification. RFC comprises a collection of decision trees, with each tree generated using a random vector drawn independently from the feature vector. During training, each tree's training set is formed from a bootstrap sample of the data, selecting frames with replacement. Each tree then casts a vote for the most probable class to predict the output. To develop and evaluate our proposed model, the classifier was trained multiple times, recording the best accuracy achieved. The frames were classified into diverse domain classes.

When EdgeMA is executed, it, by default, retrieves the last ten frames from the batch, converts them to grayscale, and extracts features to ascertain the domain. To evaluate our proposed detector, we train it multiple times and record the highest accuracy, as discussed in Section \ref{test1}.

\subsection{Label Distribution Shift Detection and Model Adaptation}

After identifying the specific domain, EdgeMA fine-tunes the model using the corresponding domain training set. Fine-tuning employs importance weighting\cite{fang2020rethinking} (IW). This powerful technique applies a weight of importance to each class in the training set based on the distributional feature, which captures similarity to the live data distribution.

\begin{figure}[htbp]
 \centering
 \includegraphics[width=0.95\textwidth]{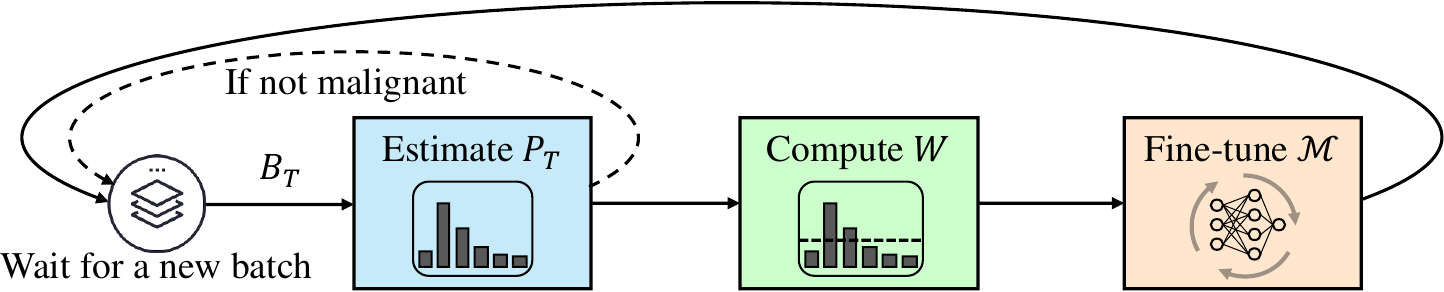}
 \caption{Streaming schema of the main operations.}
 \label{adaptive learning}
\end{figure}

The label distribution of the training set $X_S$ for the current domain is $P_S(X_S)$ (default IID). And the label distribution of the dataset $X_T$ consisting of all $N$ frames in the batch $B_T$ is $P_T(X_T)$. The distribution $P_T(X_T)$ of different windows is non-IID and dynamically varying. Figure \ref{adaptive learning} shows the streaming schema of main operations.

\textbf{Estimate $\bm{P_T}$}. The model $M$ identifies $K$ classes in total, and the predicted value of these is $f(*)$, then the distribution $P_T(X_T)$ is as follows:

\begin{equation}
P_T\left(X_T\right)=\left\{\sum_{j=1}^N \frac{\left\{f\left(x_j\right)==i\right\}}{N}\right\}_{i=1}^K   \label{e7}
\end{equation}

\noindent where $\{f(x_j)==i\}$ is a Boolean operation (1 if equal, 0 if not). The distribution outcome is a vector.

\textbf{Compute $\bm{W}$}. Similarly to \cite{lipton2018detecting}, the importance weight $W(X)$ is calculated as follows. It explains that the class is given a higher weight with a higher probability of occurrence:

\begin{equation}
W(X)=\frac{P_T(X_T)}{P_S(X_S)}=C_{h_0}^{-1} q_{h_0}   \label{e8}
\end{equation}

\noindent where $h_0$ is the initial neural network trained on $X_S$, $C_{h_0}$ is the confusion matrix of $h_0$ in source label distribution $P_{S}$, and $q_{h_0}$ is a K-dimension vector $q_{h_0}[i] = P_T(h_0(X)=y_i)$.

\textbf{Fine-tune $\bm{M}$}. Finally, model $M$ is fine-tuned with $W$ to cope with the ever-changing distribution. To reduce time cost, we utilize \textit{coordinate descent} \cite{nesterov2012efficiency, wright2015coordinate}. In this approach, EdgeMA retrains a limited subset of parameters (e.g., 20\%) during each fine-tuning phase, with the edge device refining them over \( k \) iterations. The final optimization target for fine-tuning the model is shown below:

\begin{equation}
\frac{1}{n} \sum_{i=1}^n \frac{P_T\left(y_i\right)}{P_S\left(y_i\right)} f\left(x_i, y_i\right) \rightarrow \mathbb{E}_{x, y \sim P_S}\left(\frac{P_T(y)}{P_S(y)} f(x, y)\right)   \label{e9}
\end{equation}

Notably, fine-tuning can take up computing resources on the edge device. EdgeMA needs to determine the malignancy of a shift and reduces model update frequency. EdgeMA allows the model lags \textit{if not malignant}. In practice, distributions shift constantly, and often these changes are benign. We employ the Kullback-Leibler (KL) divergence to quantify the discrepancy between the distribution $P_M$ at the final stage of fine-tuning and the current distribution $P_T$. The KL divergence is defined as:

\begin{equation}
d = D_{K L}(P_T \| P_M)=\mathbb{E}_{x \sim P_T}\left[\log \frac{P_T(X_T)}{P_M(X_M)}\right]   \label{e10}
\end{equation}

By configuring a distance threshold $D$ to determine whether adaptive learning is required, if $d<D$, it means that the model still matches the current label distribution.

\section{Evaluation}
\label{sec:evaluation}

In this section, we assess the performance of EdgeMA. Specifically, our evaluation seeks to answer two primary questions: (1) How effective is the lightweight detector, and what constitutes its peak performance? (2) How does EdgeMA cope with domain and label distribution shifts, and how does the effect of model adaptation perform in comparison to the baseline model?

\subsection{Effectiveness of Lightweight Domain Detector}    \label{test1}

We evaluate EdgeMA on the task of object detection with the UA-DETRAC\cite{wen2020ua} dataset, which contains various environments, including sunny, rainy, cloudy, and night. We first trained our lightweight detector multiple times to develop and assess it for detecting the four domain environments.

\begin{figure}[htbp]
\begin{minipage}[t]{0.5\textwidth}
\centering
\includegraphics[width=\textwidth]{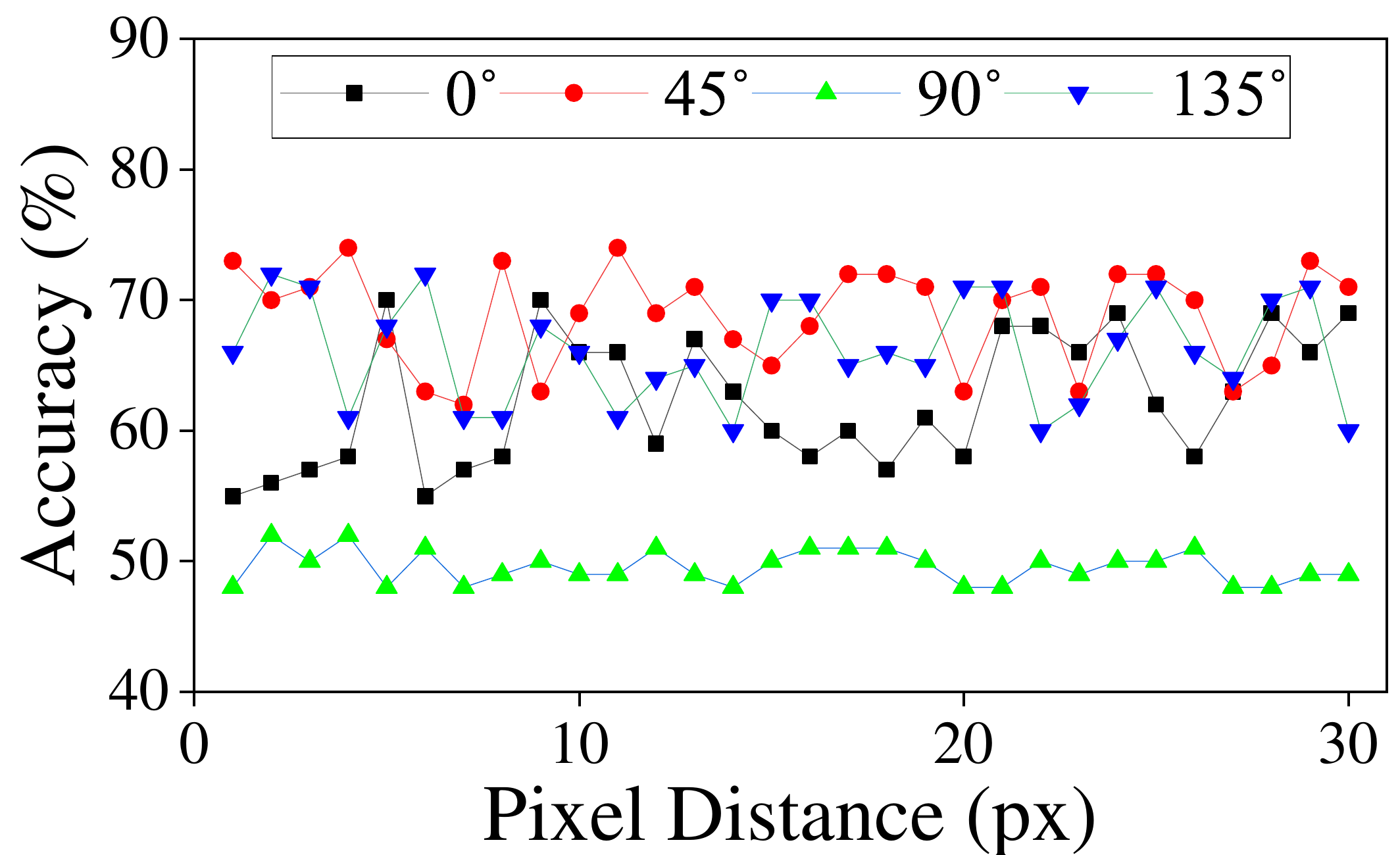}
\caption{Performance on different angles and distances.}
\label{fig3}
\end{minipage}%
\begin{minipage}[t]{0.5\textwidth}
\centering
\includegraphics[width=\textwidth]{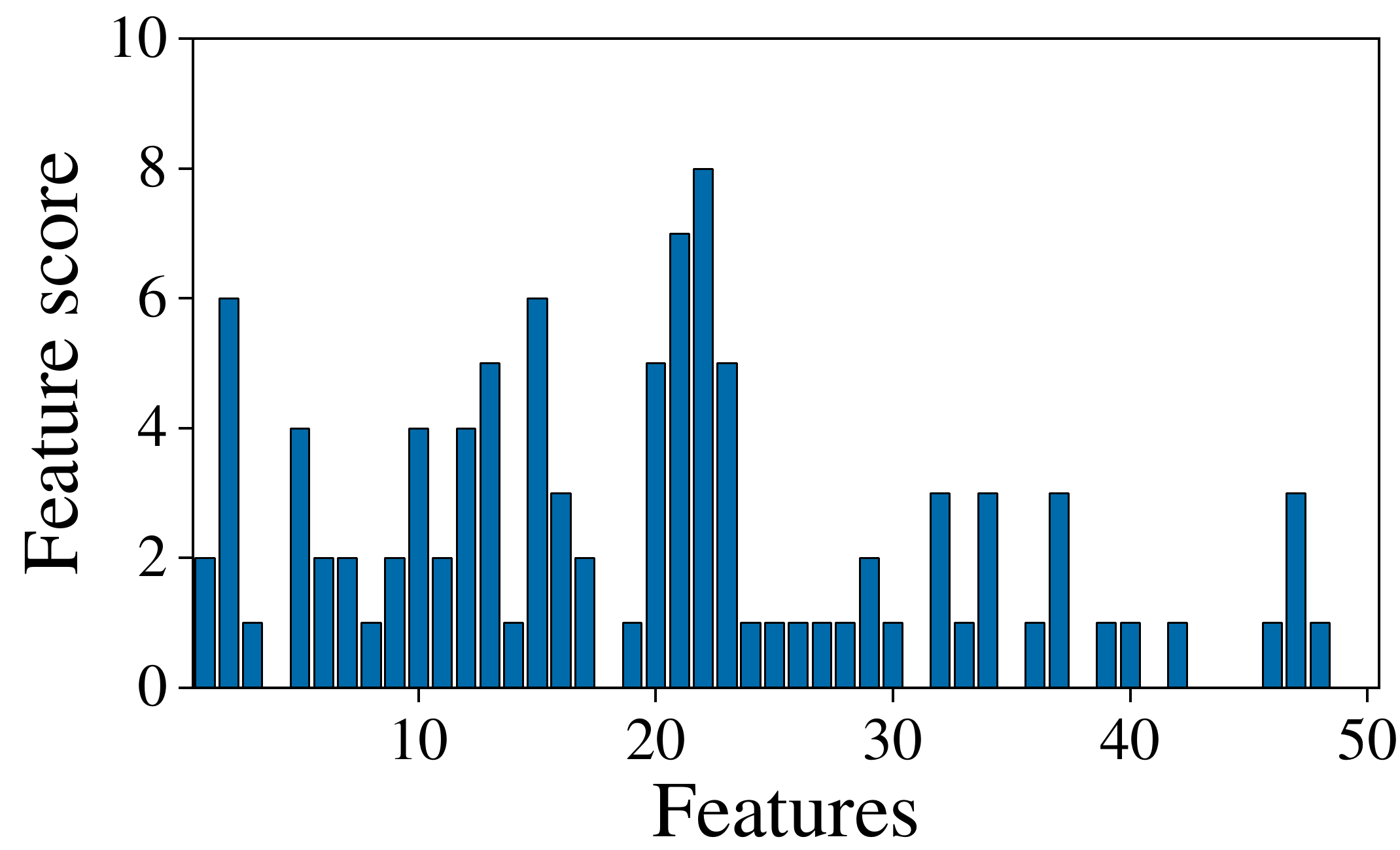}
\caption{Feature Importance.}
\label{fig4}
\end{minipage}
\begin{minipage}[t]{0.5\textwidth}
\centering
\includegraphics[width=\textwidth]{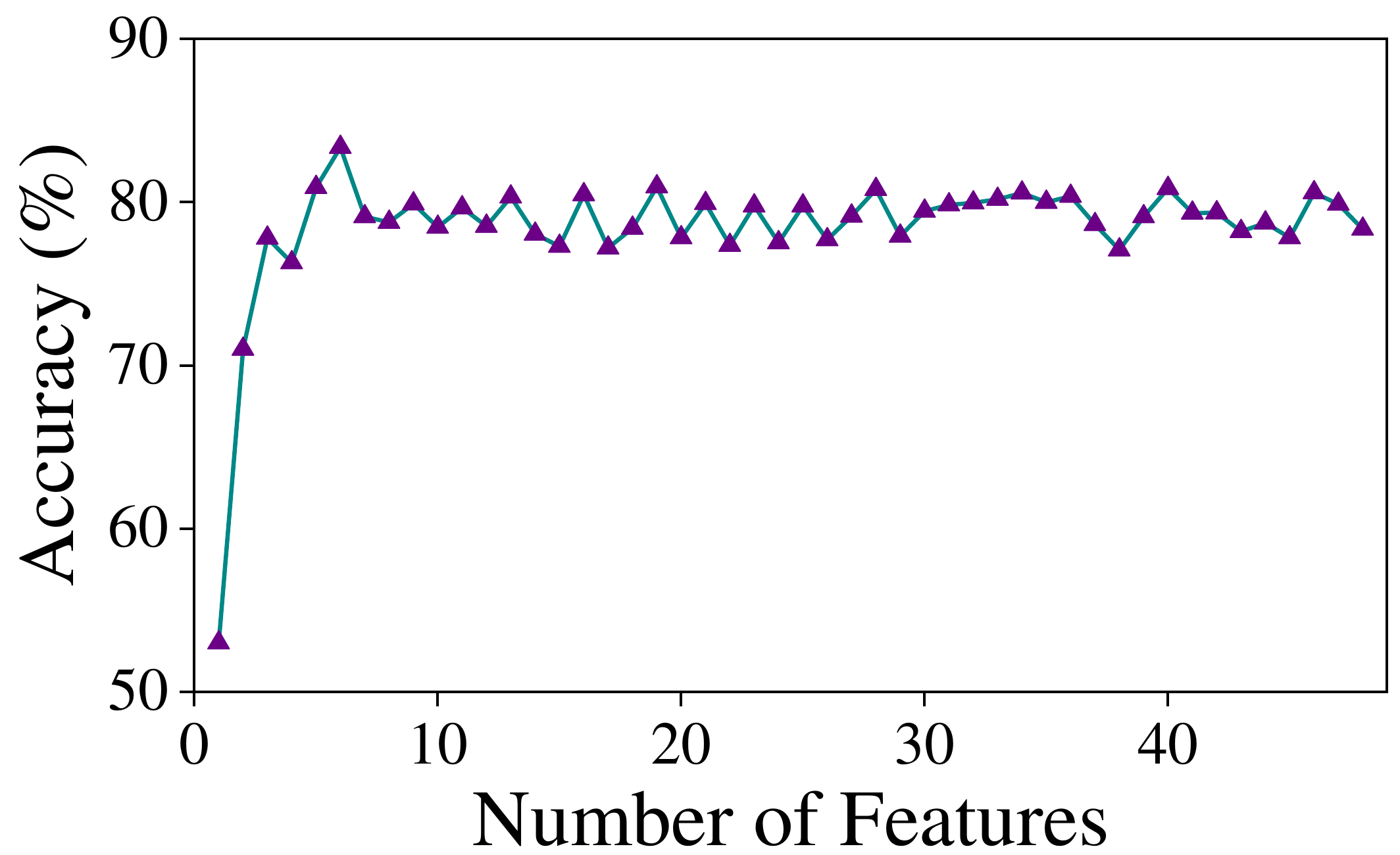}
\caption{Performance on number of best features.}
\label{fig5}
\end{minipage}%
\begin{minipage}[t]{0.5\textwidth}
\centering
\includegraphics[width=\textwidth]{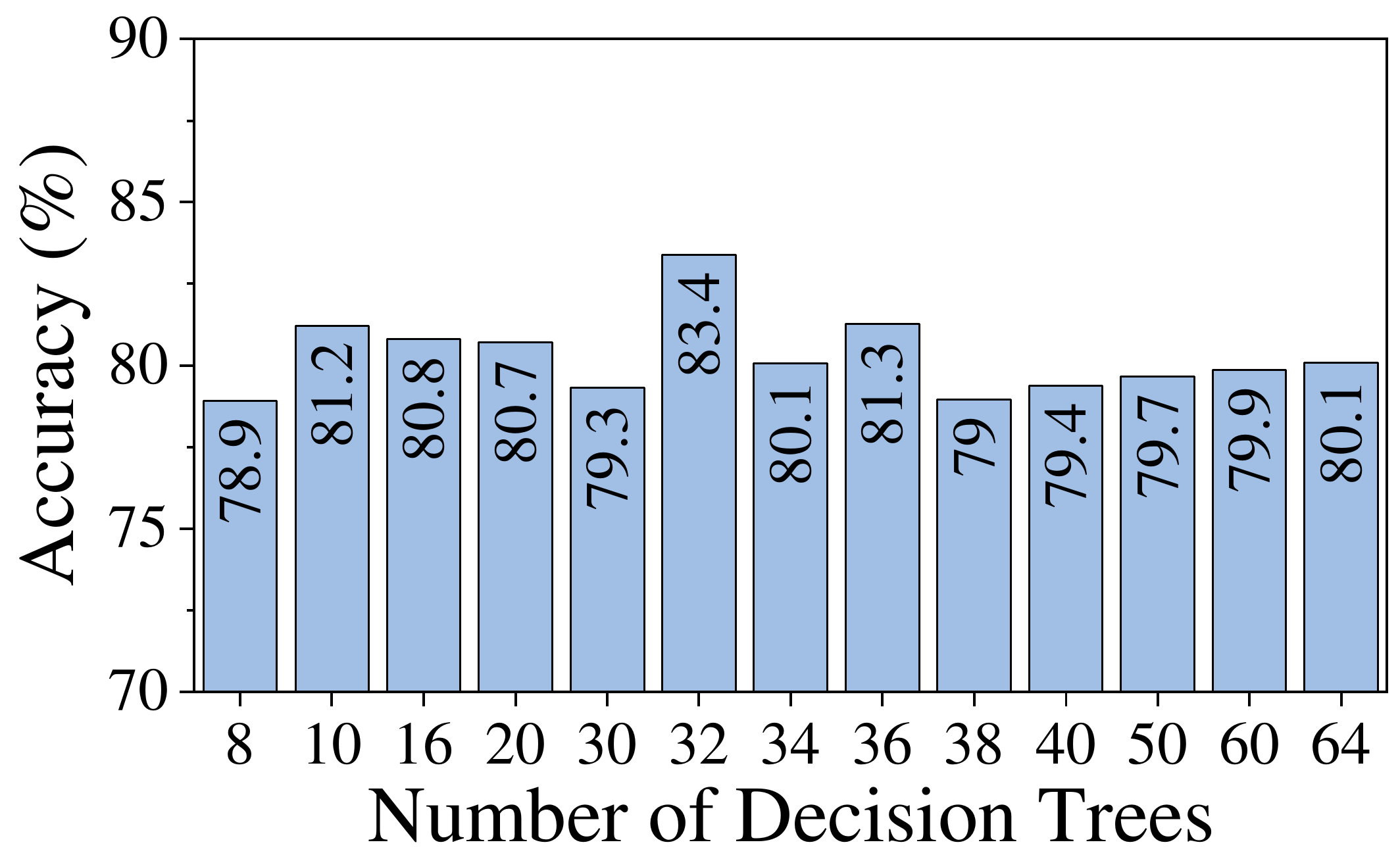}
\caption{Classifier accuracy.}
\label{fig6}
\end{minipage}
\end{figure}

Figure \ref{fig3} shows the accuracy for a collection of 6 features from GLCM using different pixel distances $d \in$ [1, 30] and different angles 0\degree, 45\degree, 90\degree, 135\degree. Then we select the 2 best neighboring pixel distances for each of the 4 angles (d=\{5,9\} for 0\degree, d=\{4,11\} for 45\degree, d=\{2,4\} for 90\degree, d=\{2,6\} for 135\degree). 
Consequently, the total number of features is calculated as 6*2*4 = 48 features.

The AdaBoost selects the top-performing features from the pool of 48 based on their feature importance. We set the iteration parameter for AdaBoost to 100. We number these 48 features. Among these, $F_{22}$ has the highest score, which denotes the correlation of the GLCM matrix with d=4 and angle 45\degree. Figure \ref{fig4} shows the importance of all features.

Then we analyze the accuracy by varying the number of top-ranked features based on their scores, as illustrated in Figure \ref{fig5}. The highest accuracy, 83.38\%, is achieved with six selected features.

We set 32 as the number of decision trees in the above evaluation. In addition, we evaluate other numbers, and the results are shown in Figure \ref{fig6}. Assigning 32 trees for the random forest still shows the best performance.

\subsection{Overall improvements of Model Adaptation}    \label{test2}
The UA-DETRAC dataset comprises over 140 thousand frames with timestamps, and we utilize these frames, concatenating part of them to form a one-hour-long video stream, 25 frames per second. We use Nvidia Jetson AGX Xavier boards as edge devices. For these devices, we use the YOLOv4\cite{bochkovskiy2020yolov4} with the Resnet18\cite{he2016deep} model backbone. We select an additional 2000 images as the training set, and we find that choosing the fine-tuning iteration k=8 is the ideal trade-off between accuracy and training time.

\begin{figure}[htbp]
\begin{minipage}[t]{0.5\textwidth}
\centering
\includegraphics[width=\textwidth]{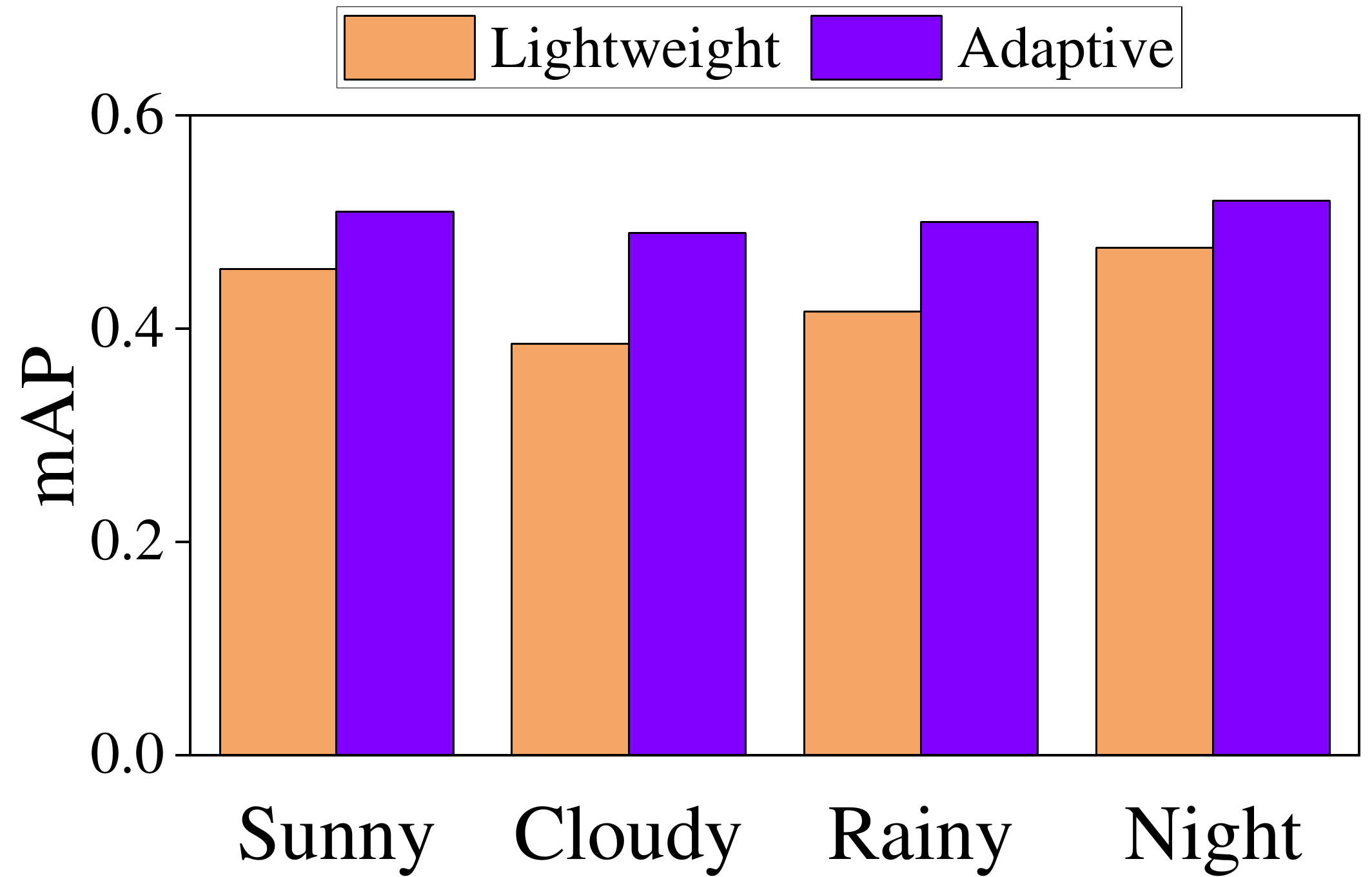}
\caption{Overall mAP.}
\label{fig7}
\end{minipage}
\begin{minipage}[t]{0.5\textwidth}
\centering
\includegraphics[width=\textwidth]{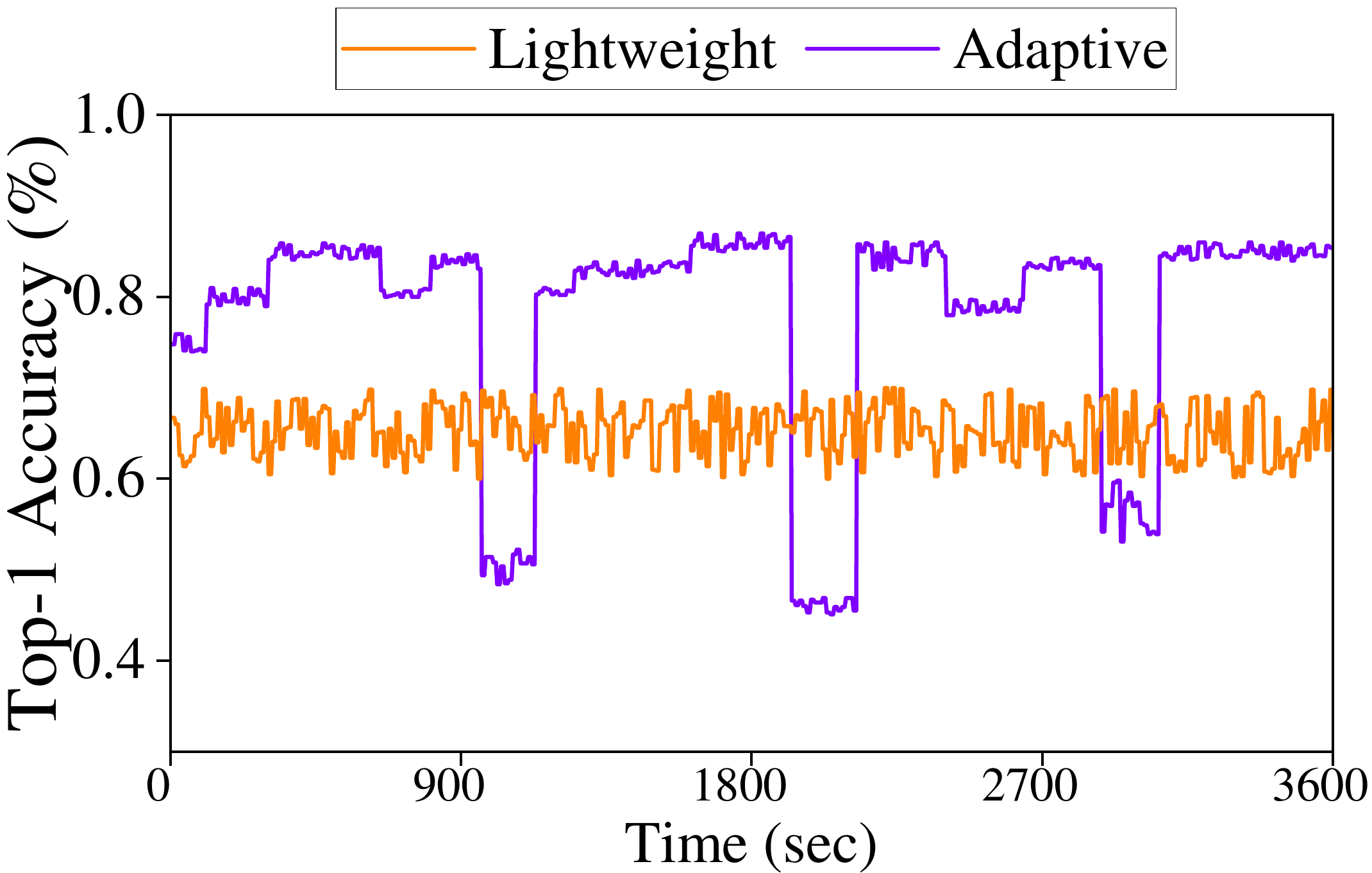}
\caption{Top-1 Acc over time.}
\label{fig8}
\end{minipage}
\end{figure}

To validate the robustness and efficacy of our model adaptation, we compare the detection accuracy (mAP metric) of the static lightweight YOLOv4 and the models fine-tuned in different domain conditions. The results are shown in Figure \ref{fig7}. The adaptative solution delivers higher detection accuracy since the model is specifically adapted to the revealed shift. Figure \ref{fig8} shows the result of the lightweight model and model adaptation on the inference top-1 accuracy of the video over time. There are several intervals when the accuracy significantly drops (about 3-5 minutes per interval) as the model lags during scene changes. Nevertheless, the average performance of model adaptation is much higher than using only a lightweight model for inference. The adaptive solution has a top-1 accuracy above 0.8 most of the time, while the fixed lightweight model solution is stable in the range of 0.6-0.7.

\begin{table}[htbp]
\caption{Comparison of two different schemes.}
\label{table:table2}
\centering
\setlength\tabcolsep{4pt}
\small
\begin{tabular}{c c c c c}
\hline
\textbf{Scheme} & {\makecell[c]{\textbf{Fraction of} \\ \textbf{Parameters}}} & \textbf{mAP@0.5} & \textbf{FPS$_{avg}$} & {\makecell[c]{\textbf{Fine-tuning} \\ \textbf{Time Cost}}}\\
\hline
\hline
\multirow{1}{*}{\textbf{Lightweight}} & - & 0.406 & 59.6 & - \\
\hline
\multirow{4}{*}{\makecell[c]{\textbf{Model} \\ \textbf{Adaptation}}} & 5\% & 0.470 & 56.3 & 95 s \\
& 10\% & 0.493 & 54.9 & 123 s \\
& \cellcolor{Gray} 20\% & \cellcolor{Gray}0.515 & \cellcolor{Gray}52.2 & \cellcolor{Gray}192 s \\
& 100\% & 0.523 & 37.5 & 887 s \\
\hline
\end{tabular}
\end{table}

Table \ref{table:table2} summarizes the performance of two different schemes. Fractions for model adaptation represent the percentage of how many model parameters for each fine-tuning. Selecting only 20\% of model parameters performs very effectively, achieving $>$0.1 better mAP score than the lightweight model. It results in only a 0.008 loss of mAP on average, but it reduces the fine-tuning time cost from 887s for full-model fine-tuning to 192s. In addition, due to the usage of computing resources for model fine-tuning, it shows a slight speed advantage over the model adaptation scheme.

\section{Conclusion}
\label{sec:conclusion}
In this paper, we propose EdgeMA, an innovative framework designed for video analytics on edge devices, which utilizes model adaptation supported by data drift detection and adaptive retraining methods. EdgeMA fine-tunes the model based on importance weighting after detecting shifts in the domain and label distribution during inference. Our evaluation shows that EdgeMA delivers overall higher accuracy compared to the static setting without model adaptation.

\subsubsection{Acknowledgements} This work is supported by the Key Research and Development Program of Guangdong Province under Grant No.2021B0101400003, the National Natural Science Foundation of China under Grant No.62072196, and the Creative Research Group Project of NSFC No.61821003.

%
%
%
%
\bibliographystyle{splncs04}
\bibliography{ref}
\end{document}